\documentclass{article}
\pdfoutput=1

\usepackage[preprint]{neurips_2026}

\usepackage[utf8]{inputenc}
\usepackage[T1]{fontenc}
\usepackage{hyperref}
\usepackage{url}
\usepackage{booktabs}
\usepackage{array}
\usepackage{amsfonts}
\usepackage{amsmath}
\usepackage{nicefrac}
\usepackage{microtype}
\usepackage{xcolor}
\usepackage{graphicx}
\usepackage{placeins}

\newcommand{\closure}{\mathcal{C}_{Q,H}}
\newcommand{\reach}{\mathcal{R}}
\newcommand{\installed}{\mathcal{I}}
\newcommand{\rsup}{r_{\mathrm{sup}}}
\newcolumntype{L}[1]{>{\raggedright\arraybackslash}p{#1}}

\makeatletter
\renewcommand{\normalsize}{%
  \@setfontsize\normalsize\@xpt\@xipt
  \abovedisplayskip      11\p@ \@plus 3\p@ \@minus 4\p@
  \abovedisplayshortskip 6\p@ \@plus 3\p@
  \belowdisplayskip      \abovedisplayskip
  \belowdisplayshortskip 8\p@ \@plus 3\p@ \@minus 3\p@
}
\normalsize
\makeatother

\title{What a World Model Represents Is Three Questions}

\author{%
  Donna Vakalis \\
  Mila -- Quebec AI Institute \\
  Universit\'e de Montr\'eal \\
  Montr\'eal, Qu\'ebec \\
  \texttt{donna.vakalis@olympian.org} \\
}

\begin{document}

\maketitle

\begin{abstract}
World models learn task-relevant information through many routes: observation
reconstruction, recurrent state, temporal filtering, and explicit task supervision.
Different routes can make different variables available. The same variable can also be available
through several routes at once. When it is, looking at which route would increase the training loss
most if removed does not tell you which route the model actually uses.
The questions are \emph{reachability}, whether a training signal can identify a task-relevant
direction; \emph{admission}, whether that direction is recoverable from the latent; and
\emph{assignment}, which eligible route carries it.
We test them in environments with a known set of required
coordinates. A direction cannot enter the latent unless some training signal can identify it. Reconstruction,
recurrence, or filtering may already recover some of those coordinates; a reward or value head then has
no residual direction to admit. For what remains, how many independent predictions the target supplies is
how many coordinates install: one through four independent predictions admit one through four directions,
including through the value head. Reachability is not admission: a temporal second-moment
coefficient can remain absent under next-token prediction when accumulating it is a fraction of a
percent of that loss, and a head that predicts the coefficient restores it. Assignment is a different test. Two routes that each carry the same variable when trained alone do
not swap the carrier when we reverse which is more costly to remove. A recurrent model trained on a transformer's recorded
sequences shows the same pattern. Near the point where the competing route
is beginning to clear the probe threshold, independent training runs disagree. What a world model represents is therefore three questions:
what information is reachable, what supervision admits, and which competing route carries it.
\end{abstract}

\section{Introduction}
\label{sec:intro}

A world model does not learn from a single undifferentiated signal. Its representation is shaped by 
several training routes, including observation reconstruction, temporal prediction, and explicit query or value heads, 
each of which can make different information available to the learning process, for encoding in the model’s state.  
Several training signals can provide information about the same underlying variable, but through different computational routes.
For instance, a static parameter may be visible in one observation, identifiable only from a time series, or absent from the
observations and supplied only as a prediction target. Treating representation as a yes-or-no property obscures this route structure.

The questions to ask instead are \emph{reachability}, \emph{admission}, and \emph{assignment}.
Reachability asks whether a training route can identify a task-relevant direction.
Admission asks whether that direction is recoverable from the trained latent above a fixed measurement threshold.
Assignment asks which of several eligible routes carries an admitted direction, as identified by
interventions on a frozen representation. A direction cannot enter the latent unless some training signal can identify it.
Admission and assignment still split after that, even in an over-parameterized model. A variable can be available
through two routes but carried by only one; a route can improve
held-out prediction weakly while failing the pre-registered probe threshold that identifies the
carrier; and the route that would reduce the training loss most if learned can still lose the
competition, a route-level analogue of feature competition in supervised networks
\citep{shah2020,hermann2020}.

The reference object is the \emph{task closure}: the predictive degrees of freedom needed to answer
a query family through a finite horizon. Closure says what a sufficient world-model state must
retain; it generalizes fixed-query sufficient states such as the information state and
action-sufficient representations \citep{subramanian2022approximate,huang2022action}.
It does not say how those directions arrive in the learned state. We factor that process into stages:
\begin{equation}
\text{channel reach}
\quad\longrightarrow\quad
\text{admission}
\quad\longrightarrow\quad
\text{assignment}.
\label{eq:stages}
\end{equation}

\begin{figure}[t]
  \centering
  \includegraphics[width=.98\linewidth]{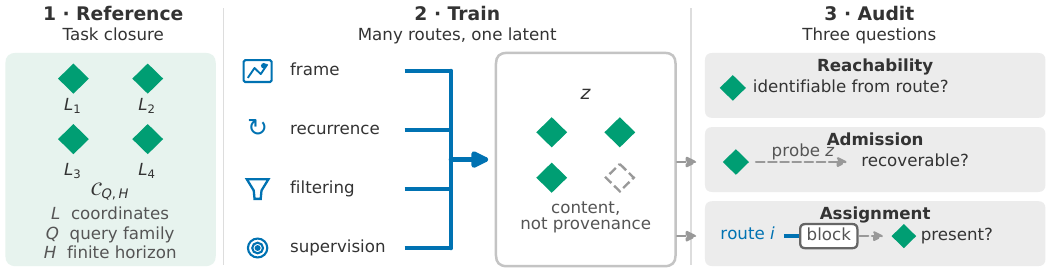}
  \caption{\textbf{Representation formation is not one decision.} Task closure specifies the
  directions a sufficient state would retain. Channel reach asks whether a training route can
  identify a direction; admission asks whether it is recoverable from the trained latent; assignment
  asks which eligible route carries an admitted direction and requires route-specific interventions.
  A scalar aggregate loss does not distinguish these estimands. (The top-to-bottom route list is
  schematic, not an ordering.)}
  \label{fig:visual-abstract}
\end{figure}

What makes this a claim about world models, rather than about any learned representation? Part of it
is not: reachability, admission, and assignment can be asked of any representation trained through
several signals, and the supervision-rank count concerns prediction targets, not dynamics. The rest
is temporal in an essential way. Remove the horizon and closure collapses to the minimal sufficient
statistic of supervised representation learning \citep{achille2018emergence}; the recurrence and
filtering routes do not exist for a static model. What we study is state formation, not state
propagation: whether the learned transition advances an admitted direction correctly is a separate
question, and a state can contain a task variable while advancing it incorrectly
\citep{vakalis2026operator}. What earns the term ``world model'' is the training recipe the modern
literature means by it: one latent shaped simultaneously by reconstruction, recurrent carry, and
reward and value heads. That multi-signal stack is not incidental: it is what creates several
training signals that can identify the same direction, and hence what makes assignment a live
question for our field.

This factorization changes how the role of the training objective should be stated. Every conventional
training objective is a scalar function. What can have rank is the family of targets or prediction
constraints aggregated inside that scalar loss. We call its effective rank the \emph{supervision
rank}. In the controlled linear construction, if the model is trained to predict
$Y=U^\top L$ for closure coordinates $L$, then
\begin{equation}
\rsup = \operatorname{rank}\!\left(\operatorname{Cov}(Y)\right)
      = \operatorname{rank}\!\left(U^\top \Sigma_L U\right).
\label{eq:supervision-rank-intro}
\end{equation}
Here $L$ is the vector of closure coordinates, $U$ is the matrix of linear combinations that form the
prediction target $Y$, $\Sigma_L=\operatorname{Cov}(L)$, and $\rsup$ is supervision rank, the number
of independent constraints in $Y$.
The loss may sum all coordinate errors into one number while $Y$ supplies several independent
constraints. Conversely, a scalar target has supervision rank one in our linear construction because
of how it is built, not because every scalar reward is intrinsically one-dimensional.

Our admission experiments sharpen the value-equivalence story. In a controlled DreamerV3 testbed, a 
latent trained on an aligned scalar value target recovers \(R^2=0.10\) of a four-dimensional closure, 
compared with \(R^2=0.765\) under the matched full target. Supervision rank counts the independent 
combinations of closure coordinates predicted by the target. Sweeping this rank from one to four, 
while the loss remains scalar, installs exactly one through four closure directions. The same 
one-for-one pattern appears through both an auxiliary prediction head and the model’s value head, 
with weaker recovery but unchanged installed rank. A family of scalar values can therefore provide 
high-rank supervision when its members constrain different combinations of the closure 
coordinates \citep{grimm2020,grimm2021}. Single-reward value equivalence is one rank-one instance, 
not a limitation of the principle.

Admission is also route-relative. When closure coordinates are visible in each frame, reconstruction
installs them, and a scalar reward is representationally sufficient. A
transient cue can instead be carried by recurrence, and a static parameter with no single-frame
signature can be admitted through temporal filtering. Intuitively, these cases correspond to an object’s 
color, visible in every frame; an instruction shown once, directly visible at first but available later 
only through memory; and friction, never directly visible but inferable from motion over time.
They establish the reachability boundary: supervision rank governs admission only for
directions not already recovered through reconstruction, recurrence, or filtering.

Reachability still does not imply admission. On a separate environment, an episode-level autoregressive
coefficient is identifiable from the lag-1 second moment of one observation channel and has no
single-frame signature. A model trained only to predict the next token does not accumulate that
statistic. An exact-law decomposition of next-token loss prices the accumulable term at about
$0.3\%$ of total next-token cross-entropy. Adding a head that predicts the coefficient, with the
same model family, data, and step budget, restores a generalizing read. The model class can
represent the statistic; next-token prediction by itself does not install it. This analysis is
exploratory and scoped to that environment.

Assignment begins where routes overlap. We construct a variable reachable through both a direct
frame channel and a harder temporal channel, measure the extra training loss when each route is
removed, and pre-register which route should carry the variable as that ordering reverses. The prediction fails.
Reconstruction carries the variable on both sides of that reversal. It wins even where the temporal
route has a large measured demand advantage. The losing route is not hypothetical: when
reconstruction is removed, the same model family, experimental setting, and frozen probe recover the
temporal carrier at $R^2=0.8969$. We knew which route should win if extra training loss selected the
carrier; it did not.
The same pattern, frame recovery above threshold and temporal recovery below it, appears when a
recurrent model is trained on the compact transformer's recorded sequences. That transfers the
ordering across architectures. Demand's failure does not hand assignment to some other property of
the training recipe. Near the point where the competing route is just beginning to clear the probe
threshold, the same architecture, data, and extra-loss ordering still install in some independent
training runs and not others.

The paper makes five contributions:
\begin{itemize}
  \item We separate representation formation into \emph{reachability} (is the information
  identifiable from training?), \emph{admission} (does it enter the latent?), and
  \emph{assignment} (which eligible route carries it?). This distinction matters because failure
  at each stage calls for a different explanation and intervention. The reference object is task
  closure with the query family left free; its effective rank makes minimal model size measurable.

  \item We define supervision rank as the number of independent prediction constraints and show,
  in our controlled regime, that one through four constraints admit one through four closure
  directions, including through the model's value head. This explains why a scalar overall loss
  can nevertheless provide high-rank supervision.

  \item On one environment, a filtering-identifiable coefficient is present in the training input and
  absent from every tested state location. An exact-law decomposition prices accumulating it at
  about $0.3\%$ of next-token cross-entropy. Adding a head that predicts the coefficient, with the
  same model family, data, and step budget, restores a generalizing read. The analysis is exploratory.

  \item We test two routes that are each independently able to carry the same variable, then
  reverse which route is more costly to remove. The carrier does not reverse, showing that
  extra training loss is insufficient to predict where information is carried.

  \item We reproduce this extra-loss versus carrier dissociation across transformer and recurrent
  stacks and find that, near the onset of threshold crossing, independent training runs disagree.
  These results separate which route carries the variable from weak leftover information or a model
  that cannot represent it, and show that assignment need not be deterministic.
\end{itemize}

Representational failures have different causes, and therefore different remedies. Unreachable
information cannot be recovered by stronger supervision. A reachable statistic can still fail to
enter the state if next-token prediction does not pay to accumulate it. In our controlled regime, a rank-deficient
target cannot be repaired by weighting the same scalar more heavily. When several routes can carry a variable, their loss
contributions do not identify the carrier. We do not claim a universal ordering of training routes,
nor a mechanism for the winning carrier. The claim is the decomposition, and its consequence:
representation cannot be engineered or diagnosed from aggregate loss alone. 

\section{Task closure, reach, admission, and assignment}
\label{sec:framework}

\subsection{Task closure is the reference object}

Let $Q$ be a family of queries, $H$ a finite horizon, and $\mu$ a distribution over states and
action sequences. The task closure $\closure$ is the smallest predictive set of coordinates needed
to answer $Q$ under the dynamics through $H$ on the support of $\mu$. In a linear system it is the
joint reachable--observable subspace of balanced realization theory \citep{moore1981}. In a
nonlinear system it is an approximately invariant family of observables containing the queries, in
the sense of Koopman analysis \citep{mezic2005,brunton2022}. The object is task-, horizon-, support-, and feature-class
relative; it is not an intrinsic dimension of the environment.

Closure separates task sufficiency from observation fidelity. A high-dimensional observation may
contain a low-dimensional closure, while a signal occupying one covariance direction may require two
predictive coordinates. A sinusoid is the elementary example: its instantaneous samples occupy one
scalar channel, but predicting it requires both phase components
(Appendix~\ref{app:cov-closure}).
Reconstruction quality therefore does not determine whether a latent contains the predictive
directions needed by a task.

We use controlled environments in which the closure coordinates are known. This is a measurement
instrument, not a natural-image benchmark. It makes route availability, target rank, and latent
readout independently controllable. The claims concern learned representation under these
constructions and the stated world-model stacks.

\subsection{A route is an information set plus a training path}

A \emph{route} has two parts. Its information set specifies which observations, histories, actions,
or labels are statistically available to identify a closure direction. Its training path specifies
how gradients induced by that information can shape the latent. We distinguish four recurring
families:
\begin{enumerate}
  \item a frame or reconstruction route, in which the direction is identifiable from an individual
  observation;
  \item a recurrent route, in which a transiently observed direction can be retained in deterministic
  state until it is needed;
  \item a filtering route, in which a direction is identifiable only from a temporal statistic or
  known input--response pair; ``filtering'' carries its estimation-theoretic sense of inferring an
  unobserved coordinate from a time series; and
  \item an explicit supervision route, in which a query, reward, value, or auxiliary target supplies
  the direction to a prediction head.
\end{enumerate}
These labels describe experimental information sets, not a universal architectural hierarchy. A
particular model may contain additional bypasses, and every route claim requires checking that every
information path into the latent has been named and, where needed, blocked.

For route $i$, let $\reach_i\subseteq\closure$ contain the directions statistically identifiable
from the observations, histories, actions, or labels that route is allowed to use under the fixed
data protocol. Population reach is defined without reference to a particular probe. An empirical
decoder supplies only a feature-class-relative lower bound on it.
For the joint training information set, data processing gives the necessary condition
\begin{equation}
  \installed \subseteq \reach_{\mathrm{joint}},
\label{eq:reach-necessity}
\end{equation}
where $\installed$ is the set of admitted closure directions and $\reach_{\mathrm{joint}}$ is the set
identifiable from the joint training information. The inclusion holds
provided that the registered channels exhaust the trainer's inputs and exogenous optimization
randomness is independent of the data-generating state. Per-route reaches need not add to the joint
reach in general: two channels may be uninformative separately but informative jointly. Our
constructions either use nested channel information sets or certify the relevant joint reach
directly.

\subsection{Admission and weak information}

A closure direction is \emph{admitted} when a frozen, low-capacity probe recovers it from the trained
latent above a registered threshold on held-out data. Admission is deliberately instrument-relative.
The threshold defines the operational content of an installed-rank statement; it does not turn a
below-threshold read into a claim of zero mutual information.

We report weak channels separately. A route may remain below the linear decode threshold while
improving held-out likelihood, or it may be visible to a nonlinear probe but not to the registered
linear instrument. That is a \emph{decode null}, not an information-free route. The recurrent-stack
experiments in Section~\ref{sec:assignment} depend on this distinction, as does the
filtering-identifiable coefficient in Section~\ref{sec:theta4-reach-admit}: absence from every
tested state read is failure under this reader, and the later supervision control shows that
information sufficient to build the accumulator was present in the input.

\subsection{Supervision rank}

The phrase ``dimension of the objective'' is misleading because the optimized loss is scalar. The
relevant object is the rank of the prediction constraints assembled inside that loss. In the linear
construction,
\begin{equation}
  \rsup := \operatorname{rank}(U^\top\Sigma_LU).
\label{eq:supervision-rank}
\end{equation}
$L\in\mathbb{R}^k$ is the closure-coordinate vector, $\Sigma_L$ its covariance, $U$ the matrix of
linear combinations that form the target $Y=U^\top L$, and $\rsup$ the resulting supervision rank.
This definition is invariant to the factor used to whiten $\Sigma_L$ and automatically drops
directions absent from the training distribution. When $U^\top\Sigma_LU$ is nonsingular,
$\rsup$ equals the target width; otherwise target width is only a nominal count.

The nonlinear extension is more delicate. The effective rank of a whitened expected-gradient outer
product can measure how many local covector directions a nonlinear target supplies, but it depends on
the state chart and can miss non-local query structure. A nonlinear scalar reward can have gradients
spanning several state directions even though its output is scalar. We therefore make the admission
claim for the controlled linear target family and treat nonlinear supervision rank as a measurement
problem, not a settled invariant.

\subsection{Assignment requires interventions}

Admission says that a direction is recoverable from the latent; it does not identify its carrier.
Assignment is identified by intervening on one frozen trained representation: remove one information
stream at a time, with the drop/preserve/floor/placebo table and in-support checks in
Appendix~\ref{app:carriage}. The probe returns frame, temporal route, both, neither, or instrument
failure rather than forcing a binary winner.

Two channels \emph{co-reach} a direction when it is identifiable from each route's allowed
information. A stronger empirical condition, \emph{both routes live}, holds when each route, trained
alone at the tested setting, clears the probe threshold with those controls. If two channels
co-reach, a failure of extra training loss to pick the carrier can be read as assignment rather than
as a missing route. If both routes live, the experiment is a competition over which route carries
the variable. Neither condition names the winning carrier.

We measure a route's demand as the extra training loss when that route's information is removed and
replaced by a cheap predictor. Demand is measured before the probe that identifies the carrier, and
its ordering is pre-registered. If demand selected the carrier, reversing which route is more costly
to remove would reverse the route identified as the carrier, once that ranking is separated beyond
measurement error. Section~\ref{sec:assignment} tests that implication.

\section{Controlled measurement of learned content}
\label{sec:measurement}

\subsection{Latent-recovery testbed}

\begin{figure}[t]
  \centering
  \includegraphics[width=.96\linewidth]{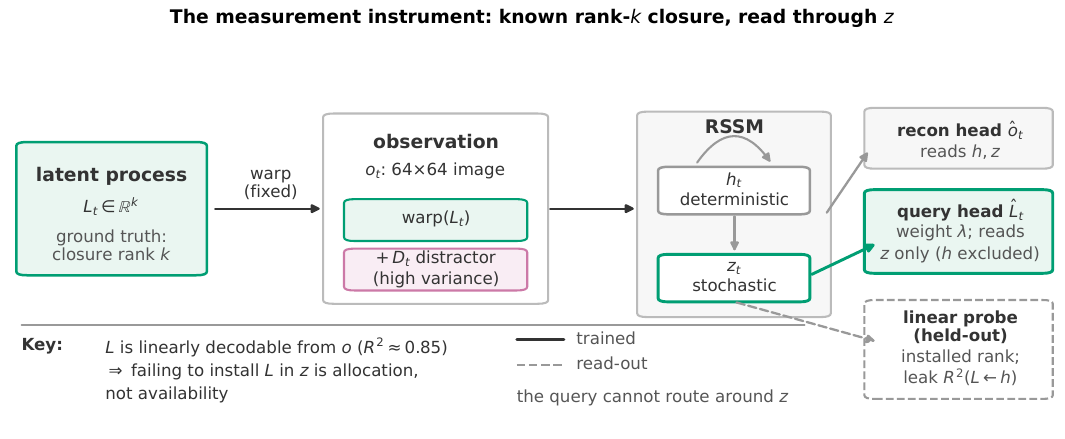}
  \caption{\textbf{Admission instrument.} Known closure coordinates and a reconstruction-salient
  distractor enter a controlled observation process. Reconstruction reads the recurrent and
  stochastic state, whereas the query head is forced through $z$; frozen held-out probes read the
  content installed in $z$, with $h$ reported to check that the query is not being satisfied
  through the recurrent state.}
  \label{fig:admission-instrument}
\end{figure}

\begin{figure}[t]
  \centering
  \includegraphics[width=.96\linewidth]{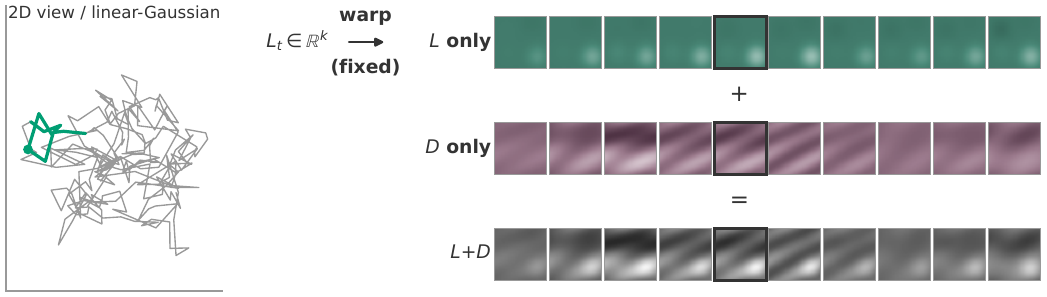}
  \caption{\textbf{Admission observations.} A known $k$-dimensional closure is
  mapped by a fixed warp into a $64{\times}64$ image. Left: a two-dimensional
  view of the $k$-dimensional linear-Gaussian trajectory $L_t$; the highlighted segment is
  warped into $L$-only frames. Right: $D$ and $L{+}D$ strips share those
  times. The marked column is one shared timestep ($L$ plus $D$). The
  high-variance distractor dominates reconstruction salience; the closure
  remains linearly readable from the same image at approximately $R^2=0.85$.
  This is a measurement construction, not a natural-image task.}
  \label{fig:admission-obs}
\end{figure}

The admission experiments use a DreamerV3-style categorical recurrent state-space model
\citep{hafner2023} (Figure~\ref{fig:admission-instrument}). A known process of $k$ slowly varying
closure coordinates is mapped through a fixed analytic warp into a high-dimensional observation
alongside a high-variance distractor (Figure~\ref{fig:admission-obs}). The closure coordinates are
linearly recoverable from the observation at approximately $R^2=0.85$.
Their absence from the trained latent therefore reflects allocation by the training system rather
than absence from the data.

The model state contains deterministic recurrent state $h_t$ and stochastic state $z_t$. The main
query head reads $z_t$ alone. This architectural restriction matters: when the head may read
$[h_t,z_t]$, it can satisfy the query through $h_t$ while leaving $z_t$ occupied by the distractor.
Before running the supervision-rank sweep, we verified that forcing the head onto $z_t$ recovers the
closure in a no-distractor positive control. A label-shuffle intervention then removes target
alignment while preserving the data, optimizer, and probe.

\FloatBarrier

\subsection{Installed rank}

For each closure coordinate $L_i$, a ridge probe is fit on frozen model states and evaluated on a
held-out split. Its score is the gap in $R^2$ relative to reconstruction-only and shuffled-target
baselines. A coordinate is counted as admitted when that gap clears a per-column threshold anchored
to those baselines. Installed rank is the number of admitted coordinates. We also report $R^2$ for
the full closure, the individual coordinate scores, how the installed count moves if the threshold
is varied, and the corresponding read from $h$.

This is a deliberately low-capacity, frozen-representation read. It does not fine-tune the encoder
and cannot create the content it measures. Nonlinear probes and held-out likelihood are sensitivity
analyses, not substitutes for the frozen probes used to count admitted coordinates or to identify
the carrier. Adaptive readouts
can change a representation while ostensibly measuring it \citep{interno2026}.

\subsection{Route interventions and identifying the carrier}

The assignment experiments use controlled dynamical channels in which an episode-level parameter
$\theta$ can be made available through a direct frame statistic, a known-input temporal statistic,
or both. Route interventions substitute the relevant substream across episodes. The substitution
checks, placebo, floor, and single-route calibrations are in Appendix~\ref{app:carriage}. Demand is
the extra training loss when a route's information is removed and replaced by a cheap predictor.
Analytic values are checked against empirical estimates. We choose two settings so that which route
is more costly to remove reverses, then inspect which route the probe identifies as the carrier. An
independent training run, rather than a checkpoint or a probe site inside a run, is the dependence
unit throughout.

The experiments span a DreamerV3-style stack, a compact transformer world model, and a recurrent
categorical RSSM. Cross-architecture comparisons train the recurrent model on the transformer's
recorded sequences under a matched protocol; they transfer an ordering, they are not independent
replications. Each result is stated at the architecture, schedule, budget, data, and probe
threshold on which it was established.

Section~\ref{sec:theta4-reach-admit} uses a further categorical RSSM on a different synthetic
environment, in which an episode-level autoregressive coefficient $\theta_4$ is identifiable only from a
temporal second moment. The single-frame render of that coefficient is removed during training, so
only the temporal statistic is available. That environment is not the assignment competition of
Section~\ref{sec:assignment}. The measurements there are exploratory: they follow a pre-registered
passive-training evaluation that failed to recover $\theta_4$ from the state, and they are scoped
to this environment, training protocol, and the time at which the state is read. A below-threshold
read is not a claim of zero information.

\section{Admission: how task-relevant directions enter the latent}
\label{sec:admission}

\subsection{Query and value heads can recruit structure reconstruction leaves behind}
\label{sec:objective-install}

Reconstruction initially dominates the stochastic latent in the distractor-rich environment. A
query head allowed to read the full recurrent state does not change this allocation: the query is
satisfied through $h$, and the closure remains absent from $z$. Forcing the head onto $z$ removes
that escape. As query weight increases, full-closure $R^2$ rises from $0.547$ to $0.877$ and
saturates near the level at which the closure is linearly available in the observation, while the
distractor remains readable.

The target-shuffle control identifies this as an effect of aligned supervision. In the
$k=2$ full-target condition, shuffling targets across training examples reduces the closure $R^2$ to
$-0.040$, while the distractor probe remains alive at $R^2=0.395$. Thus the query-aligned loss causes the
admission; neither a dead encoder nor an incidental change in the reconstruction path explains it.

\begin{figure}[t]
  \centering
  \includegraphics[width=.96\linewidth]{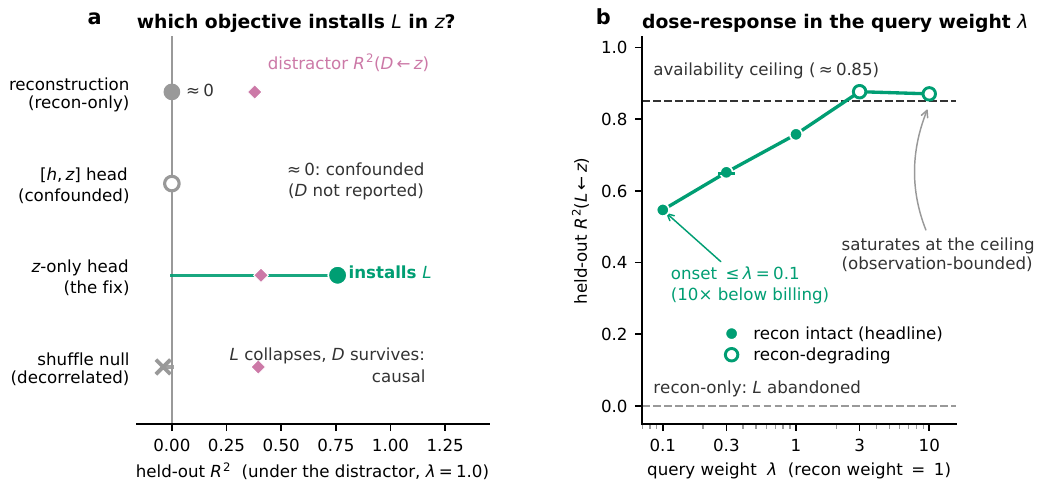}
  \caption{\textbf{A task target recruits closure directions when it is forced through the measured
  latent.} Letting the head read recurrent state allows the target to bypass $z$. Constraining the
  head to $z$ produces closure recovery that increases with query weight, while shuffling the target collapses the
  effect and leaves the distractor probe active.}
  \label{fig:objective-install}
\end{figure}

\subsection{A rank-one value target installs one projection}
\label{sec:rank-one}

We next align a scalar reward/value target with a four-dimensional closure. The target is a known
linear projection of the closure and therefore has $\rsup=1$. At the strongest registered weight,
the latent recovers $R^2=0.099$ of the full closure, below the $R^2=0.372$ bar obtained from
reconstruction-only recovery plus a pre-committed margin, while recovering its own supervised
projection at $R^2=0.487$. A placebo reward aligned with
the distractor installs the distractor at $R^2=0.874$, showing that the value head and readout are live.

In the matched four-dimensional condition, replacing the scalar target with the full closure target
raises full-closure $R^2$ to $0.765$. The model, latent size, training budget, and probe are held
fixed. Within this pair, target encoding and head form also differ, so the comparison alone is
consistent with a supervision-rank account rather than decisive evidence for it. The within-head
sweeps below remove that ambiguity.

\begin{figure}[t]
  \centering
  \includegraphics[width=.96\linewidth]{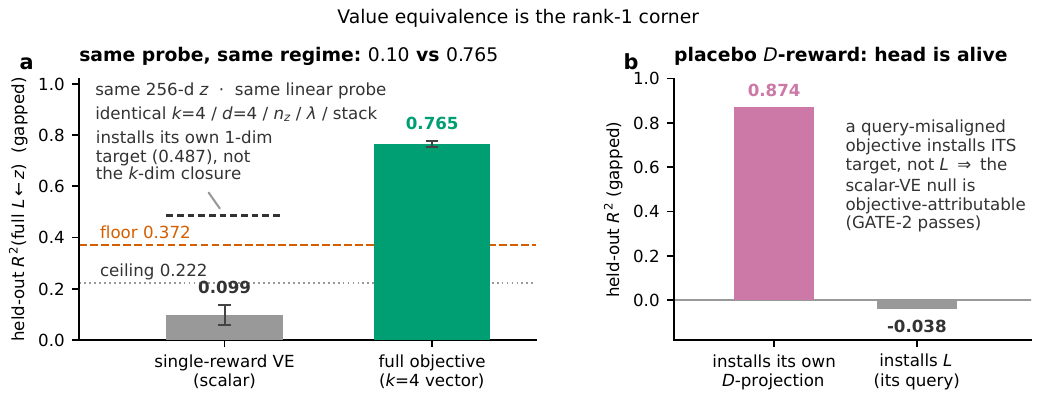}
  \caption{\textbf{The rank-one value instance.} An aligned scalar target installs its own projection
  but little of the full four-dimensional closure. A matched full target installs substantially more
  closure structure, and a distractor-aligned placebo shows that the value head can install its own
  target.}
  \label{fig:rank-one}
\end{figure}

The result does not refute value equivalence. The value-equivalence principle is defined relative to
a set of value functions, and proper value equivalence can use a high-rank family
\citep{grimm2020,grimm2021}. Our scalar construction occupies one corner of that family, the corner
that value-gradient model losses target in practice \citep{voelcker2022vagram}. Its shortcoming
follows from the single independent constraint it supplies in this environment, not from the fact
that the optimized loss is scalar.

\subsection{Installed rank tracks supervision rank}
\label{sec:supervision-step}

We vary the target family $Y=U^\top L$ from $\rsup=1$ to $4$ while holding the closure, model, and
training protocol fixed. The installed ranks are $1,2,3,4$. A weight-matched rank-one control raises
the scalar target weight without adding a target direction and still installs one direction. Shuffled
targets collapse every rank. The staircase is stable across the registered threshold band and across
the replicated seeds of the auxiliary-head experiment.

\begin{figure}[t]
  \centering
  \includegraphics[width=.96\linewidth]{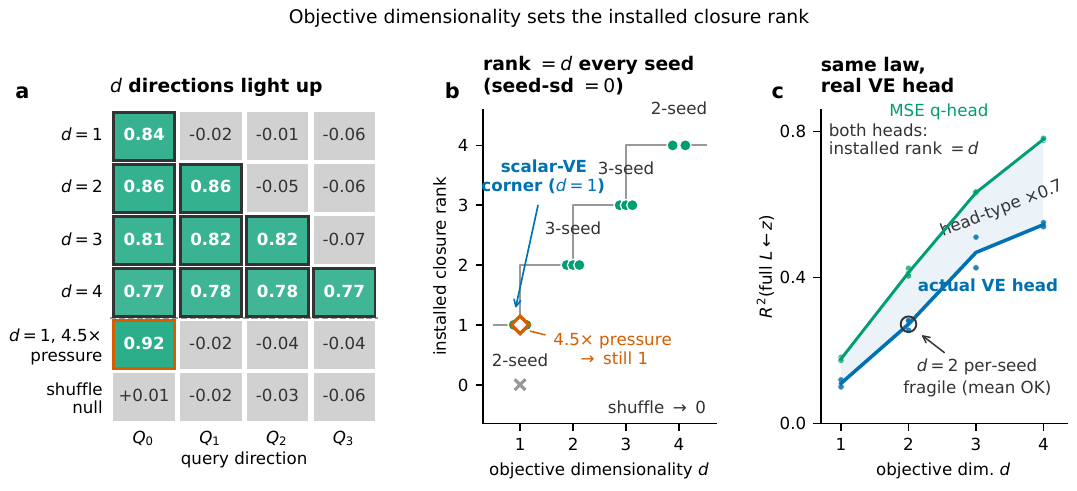}
  \caption{\textbf{Independent prediction constraints govern which remaining directions enter the latent.} Increasing
  supervision rank from one to four installs one through four closure directions. Raising the weight
  of a rank-one target does not reproduce the rank-four result; target shuffling collapses the
  staircase. The figure uses the original target-dimension label; throughout the text we call the
  measured object supervision rank.}
  \label{fig:supervision-step}
\end{figure}

The same pattern appears through the model's value head. On the three-seed mean, the installed ranks
are again $1,2,3,4$, and a rotation-invariant $R^2$ of the full closure rises
$0.110\rightarrow0.272\rightarrow0.469\rightarrow0.545$. Those $R^2$ magnitudes are approximately
$0.70$--$0.76$ of the corresponding auxiliary-head values, but rank is preserved. One seed at target rank two has
a closure coordinate the target does not supervise, sitting in the reconstruction-only null
distribution, that crosses some threshold conventions. Its sign reverses
across seeds, and the three-seed mean, the full-closure $R^2$, and all supervised columns preserve
the staircase. The claim is therefore about mean installed rank and rotation-invariant closure
recovery, not an exact per-seed count at one threshold.

The value-head sweep also reveals that the attempted loss normalization did not hold total pressure
constant: the summed value-head loss increases with target rank. Three observations keep this
from explaining the count. Every supervised column lies far above threshold even in the weakest
condition; unsupervised columns remain in the reconstruction-only null distribution; and the
separate weight-matched rank-one control does not add directions. The extra value-head loss limits
magnitude comparisons, but it does
not reproduce the rank staircase.

\subsection{Why a rank ceiling appears}

A linear target $Y=U^\top L$ can constrain at most $\rsup$ directions of $z$
\citep{anderson1951,eckart1936,izenman1975}. That bound is not a new theorem. Translating it into
installed rank needs the empirical premise that unsupported directions stay at the reconstruction
null, which the reconstruction-only, unsupervised-column, and shuffle controls support. For the
nonlinear heads used here, equality with supervision rank is an empirical law, not a consequence of
the linear geometry. Once latent width is sufficient, capacity does not set the observed plateau;
the width calibration and a trained high-dimensional-observation run that we do not treat as
pixel-scale evidence are in Appendix~\ref{app:rank-ceiling}.

\subsection{Other routes bound where supervision matters}
\label{sec:admission-boundary}

\begin{figure}[t]
  \centering
  \includegraphics[width=.96\linewidth]{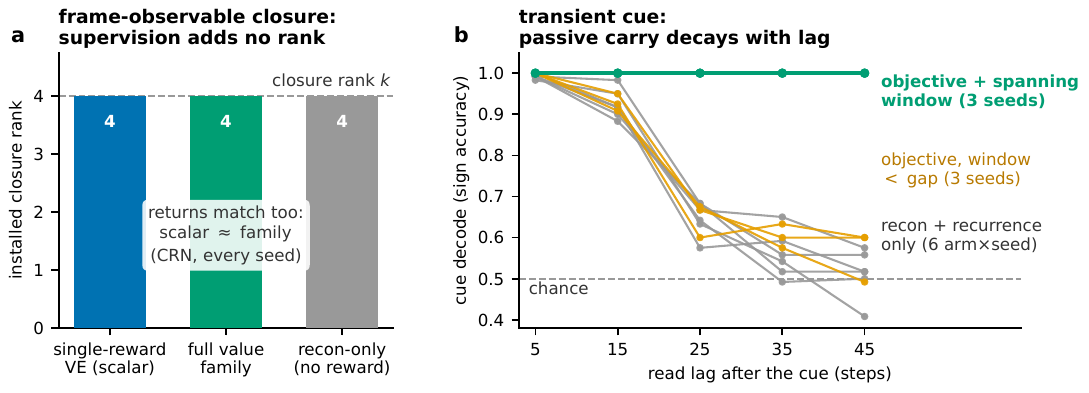}
  \caption{\textbf{Supervision acts on leftovers.} (a) When four closure coordinates are visible
  in each frame, reconstruction-only, scalar-value, and full-value-family models all admit rank
  four; scalar-value return matches the full family under a common random number (CRN) on every paired
  seed. (b) A cue visible early and absent at use is retained by recurrence without the
  objective, and that passive carry decays with lag. The text also reports a static parameter
  recovered only from a temporal statistic; it is not plotted. These are construction-specific,
  not a universal route order.}
  \label{fig:route-boundary}
\end{figure}

The supervision-rank law concerns residual directions, not the complete contents of a world-model
state. In a closed-loop control environment whose four closure coordinates are directly visible in
each observation, a single-frame MLP decodes every coordinate at approximately $R^2=0.998$.
Reconstruction-only, scalar-value, and full-value-family models all admit rank four; their posterior,
prior, and open-loop reads agree within noise, and the scalar-value model attains at least 90\% of the
full-family return in every paired seed. Here reconstruction already supplies the closure and a reward or value head
adds no representational rank (Figure~\ref{fig:route-boundary}a).

A partially observed variant gives recurrence a distinct role. A cue visible early in an episode but
absent at the point of use is retained in recurrent state independently of the objective, with passive
carry decaying as the lag grows (Figure~\ref{fig:route-boundary}b).

A further construction makes a static parameter invisible in every single frame and in the recurrence
inputs individually, while leaving it identifiable from temporal second moments. A reward-free model
admits that parameter through filtering. Recovery is positive from the temporal statistic, and
falls to the no-information baseline when the action stream is masked or crossed. That construction
is a different environment from the filtering-identifiable coefficient in
Section~\ref{sec:theta4-reach-admit}, where the temporal statistic is reachable and next-token
prediction does not install it. The static-parameter setting is studied in its own right as hidden-parameter MDPs and
online system identification, where a parameter such as friction is inferred from the response to
motion \citep{doshivelez2016hidden,kumar2021rma}; here it serves as one registered route among
several rather than as the estimation target. These are route-specific cases, not evidence for a
universal ordering among reconstruction, recurrence, filtering, and supervision.

Data quantity is another admission boundary. In the same filtering construction, installation
survives down to at least $410$ views per underlying frame; poorer settings are unpowered rather
than negative (Appendix~\ref{app:data-regime}).

\subsection{Reachability is not admission}
\label{sec:theta4-reach-admit}

\begin{figure}[t]
  \centering
  \includegraphics[width=.98\linewidth]{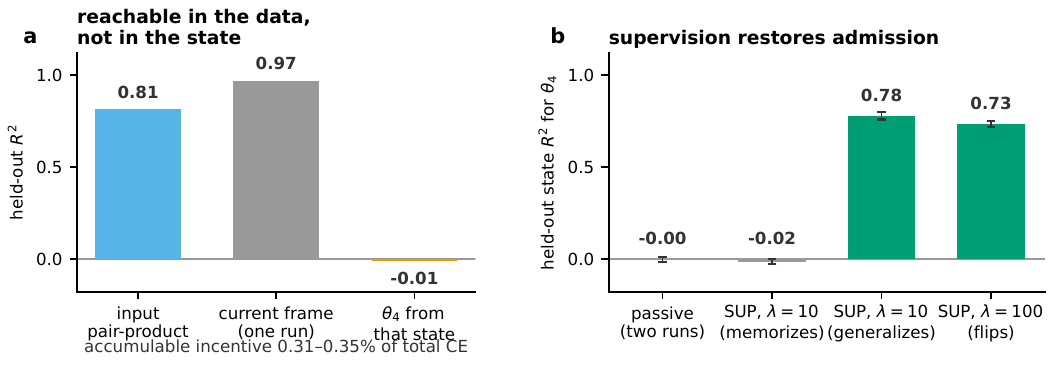}
  \caption{\textbf{Reachability is not admission.} An episode-level autoregressive coefficient
  $\theta_4$ is invisible in every single frame and identifiable from a lag-1 second moment.
  (a)~Ridge on products of successive frames recovers it at held-out $R^2=+0.81$. On one run that keeps the
  corresponding observation channel, the current frame is in the state at $R^2=0.97$ while
  $\theta_4$ from that state is at the shuffle-control level. An exact-law decomposition prices accumulating
  $\theta_4$ at $0.32$--$0.35\%$ of total training cross-entropy. (b)~A $\theta_4$ supervision head
  on the same architecture restores a generalizing state read ($R^2=+0.78$). At the default
  auxiliary-loss weight the other seed memorizes; a tenfold weight increase flips that seed to an estimator.
  Exploratory, one environment. Failure under this reader is not proof of zero information. This is not
  assignment evidence.}
  \label{fig:theta4-reach-admit}
\end{figure}

The filtering construction above is an environment on which a reward-free model does admit a static parameter
from a temporal statistic. A different environment isolates the opposite case
(Figure~\ref{fig:theta4-reach-admit}).

Let $\theta_4$ be an episode-level autoregressive coefficient. By construction it is invisible in
every single frame and identifiable from the lag-1 second moment of one observation channel. The
competing single-frame render of that coefficient is removed during training, so the only available
identifying route is the temporal statistic. This is not the assignment parameter of
Section~\ref{sec:assignment}. The measurements are exploratory: they were obtained after a
pre-registered passive-training evaluation on this environment failed to recover $\theta_4$ from the
state, and they are scoped to this environment, training protocol, and the time at which the state
is read (Appendix~\ref{app:theta4}).

$\theta_4$ is recoverable from episode-level input statistics. A ridge on products of successive
frames of the observation channel reads it at held-out $R^2=+0.81$ on both a four-parameter version
of the environment and a version that contains only $\theta_4$; a lag-1 autoregressive statistic
reaches correlation $0.88$ with $\theta_4$ by the evaluation time. Linear probes on individual
frames read approximately $0$, as the environment requires.

It is not recoverable from any tested model state. Linear and held-out MLP reads sit within
shuffle-control bands of approximately $\pm 0.02$ at encoder, recurrent state, stochastic state, and
the concatenated state, at every tested time through the episode, on both versions. Control probes
that recover a known linear functional of the same state read $0.85$--$1.0$ beside every null. The
result is failure under this reader and threshold, not proof of zero information.

Where that observation channel is present in the state at all, what is present is the current frame.
On one four-parameter run, encoder, stochastic state, and concatenated state recover the current
frame at $R^2=0.95$--$0.99$ at every tested time, while the accumulated autoregressive statistic
stays at the shuffle-control level. Other passive runs do not keep that frame either. None of
the five measured passive models admits $\theta_4$. Under this training protocol the model can
retain instantaneous input content without accumulating the temporal second-moment statistic needed
to recover $\theta_4$.

Next-token cross-entropy supplies almost no incentive to accumulate that statistic. An exact-law
decomposition of next-token cross-entropy on the relevant channel compares a current-frame-only
predictor with a predictor that also uses the exact posterior over $\theta_4$ from the quantized
history. The gap is $0.041$--$0.045$ nats, or $0.32$--$0.35\%$ of total training cross-entropy, on
every passive model in both versions. Direct supervision restores admission. With a $\theta_4$
head attached to the same architecture, data, and step budget, an independent cross-fitted state
read at the evaluation time recovers held-out $R^2=+0.778\pm 0.022$ (shuffle $-0.03$) on one seed,
and the held-out head itself reads $+0.576$. Prediction cross-entropy stays in the same range as the
passive runs ($1.80$--$1.87$ versus $1.81$--$1.84$). At the default auxiliary-loss weight the
outcome is seed-bistable: the other seed memorizes training episodes (held-out head $-0.51$; state
$-0.015$) and becomes a generalizing estimator under a tenfold increase in that weight (head
$+0.54$; state $+0.73$), paying $+0.014$ nats of cross-entropy. The same model family, width, data,
and step budget can therefore accumulate $\theta_4$ when the loss also contains a $\theta_4$
prediction target. Next-token prediction by itself does not install it.

This is not assignment evidence. Only one identifying route was available. The experiment asks whether a
reachable statistic enters the state, not which of two available routes carries it. It is also not the
supervised temporal-route control on the assignment environment in
Section~\ref{sec:cross-architecture}.

\section{Assignment: which route carries an admitted direction?}
\label{sec:assignment}

\subsection{When two routes can carry, extra training loss does not pick the carrier}
\label{sec:demand-competition}

The admission experiments isolate directions that only a query or value head supplies. Assignment
requires the opposite construction: the same direction must be available through at least two
routes. We use an episode-level plant parameter $\theta$ with a direct reconstruction route and a
harder temporal route based on a known excitation--response relation. Each route can be removed by a
selective substitution. Both routes admit $\theta$ when trained alone at the two competition settings, and
removing both routes returns the probe to its no-information baseline. The probe reports the direct
route, the temporal route, both, neither, or an instrument failure, and is checked on models trained
with known single routes.

For each route, we measure the extra training loss when its information is removed and replaced by
a cheap predictor. The two quantities share units and are checked analytically and
empirically. We choose two settings so that which route is more costly to remove reverses. Before
training, we committed the prediction that the route identified as the carrier would reverse with it.

\begin{figure}[t]
  \centering
  \includegraphics[width=.98\linewidth]{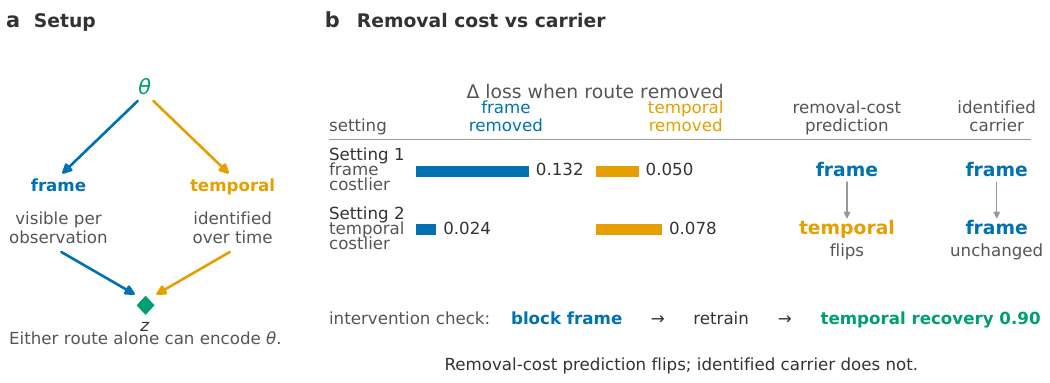}
  \caption{\textbf{Extra training loss does not identify the carrier.} An episode-level parameter
  $\theta$ is reachable through a direct frame route and a harder temporal route. Each route
  installs when trained alone. Demand is the extra training loss when that route is removed. We
  reverse which route is more costly to remove and pre-register that the carrier reverses. It does
  not: the frame route carries in both settings. In the temporal-costlier setting the identified
  carrier is separated from the extra-loss ranking by approximately 51 standard errors. When
  the frame route is removed in that setting, the same frozen linear probe recovers $\theta$ from the
  remaining temporal channel at $R^2=0.8969$, above the registered $R^2$ threshold. This is a counterexample to ranking carriers by this extra-loss comparison at the
  tested settings. It does not name the winning mechanism.}
  \label{fig:assignment}
\end{figure}

The prediction fails (Figure~\ref{fig:assignment}). The direct reconstruction route carries $\theta$ in both settings, in
every independent training run and at every pre-registered evaluation time. In the setting where
removing the temporal route increases training loss more, the separation between that loss
advantage and the observed direct-route carrier is approximately $51$ standard errors. There are no
failed controls.

The losing route is nevertheless capable of carrying the variable. When reconstruction is removed
during training in the same setting, the frozen linear probe recovers $\theta$ at $R^2=0.8969$,
above the registered $R^2$ threshold.
The temporal route therefore reaches the model, installs by itself, and is detected by the same
probe that reports its loss under competition. Conversely, removing the temporal route leaves
the direct carrier. We knew which route should win if extra training loss selected the carrier; it
did not.

This is a counterexample to any rule that picks the carrier by ranking routes on this extra training
loss, once the ranking is separated beyond measurement error. It does not refute selectors based on
integrated optimization history such as gradient starvation \citep{pezeshki2021}, initialization
alignment \citep{saxe2014}, latency, availability \citep{hermann2020,hermann2024}, or entrenchment.
Those accounts make different predictions and remain candidates for the observed carrier.

\subsection{The extra-loss versus probe ordering transfers across architectures}
\label{sec:cross-architecture}

\begin{figure}[t]
  \centering
  \includegraphics[width=.98\linewidth]{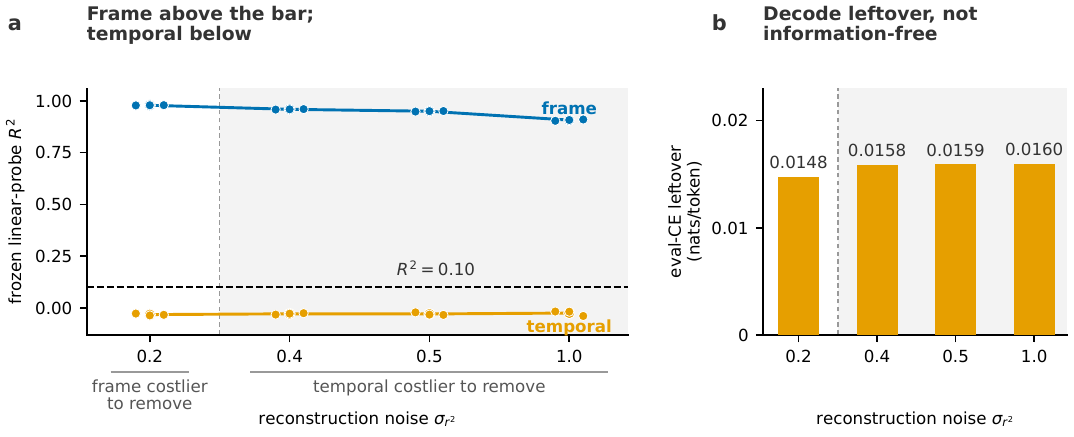}
  \caption{\textbf{The extra-loss versus probe ordering on a recurrent stack.} A categorical
  RSSM trained on the compact transformer's recorded sequences, matched step
  budget. Across four reconstruction-noise settings, frozen linear-probe $R^2$
  of $\theta$ from the frame route stays above $0.10$ and the temporal
  action--response route stays below it, including both sides of the extra-loss
  reversal. The temporal route still improves evaluation cross-entropy by
  approximately $0.014$--$0.015$ nats per token (below the probe threshold, not
  information-free). This transfers an ordering on the same sequences; it is not
  an independent replication, and it is not a competition in which each route
  carries under passive training.}
  \label{fig:rssm-transfer}
\end{figure}

A separate experiment repeats the extra-loss comparison on a categorical RSSM trained on the
compact transformer's recorded sequences, with a matched step budget. Across four settings of the
reconstruction-noise knob, the frozen linear probe recovers $\theta$ from the frame route at
$R^2=0.9784$, $0.9589$, $0.9496$, and $0.9068$.
The competing temporal, action--response route remains at
$R^2=-0.0316$, $-0.0278$, $-0.0285$, and $-0.0242$. The frame route therefore clears the registered
$R^2$ threshold at every setting while the temporal competitor remains below it, including settings on
both sides of the reversal in which route is more costly to remove (Figure~\ref{fig:rssm-transfer}).

The recurrent result transfers that ordering; it is not a second competition in which each route
was shown to carry when trained alone. The temporal variable is identifiable from the recorded
sequences, and the shared model body can represent it, but under passive training the temporal
route never cleared the $R^2$ threshold on its own. A dedicated supervised control tests whether
the shared model body can represent the variable: the RSSM recovers the temporal parameter at both
ends of the noise knob, with all six frozen linear-probe $R^2$ values above $0.920$ against the
pre-registered $R^2=0.10$ threshold, and removing the temporal stream removes the read. Recovery below threshold under
passive training therefore concerns routing under the tested protocol rather
than inability of the model or probe to represent the variable.

The null does not imply absence of information. The competing route improves evaluation
cross-entropy by approximately $0.014$--$0.015$ nats per token at every setting while failing the
frozen linear probe $R^2$ threshold. The extra-loss versus probe ordering therefore transfers across the
two architectures on the same recorded sequences. The evidence does not support ``independent
replication,'' an information-null claim, or a competition in which both routes carry under
passive training on the recurrent stack.

\subsection{Assignment is stochastic at the boundary}
\label{sec:stochastic-boundary}

\begin{figure}[t]
  \centering
  \includegraphics[width=.98\linewidth]{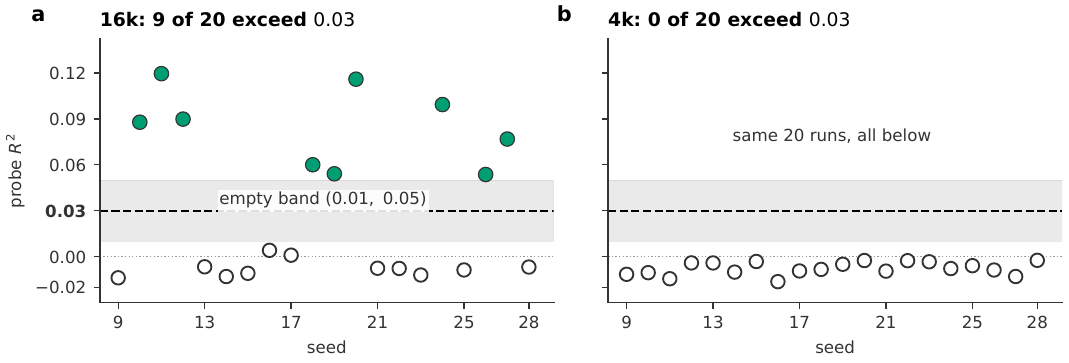}
  \caption{\textbf{Assignment at onset is a threshold-exceedance rate over runs.}
  (a)~Twenty fresh independent training runs at the registered $16{,}000$-step setting (motivating runs
  excluded): nine cross the $0.03$ probe threshold; none fall in the pre-registered empty
  band. (b)~The same twenty runs at 4k: zero crossings. This rules out a selector that is
  deterministic given architecture, data, schedule, and budget. It does not identify basins
  or a predictive initialization statistic. The $95\%$ Clopper--Pearson interval for the
  exceedance rate is $[0.2306,0.6847]$.}
  \label{fig:stochastic-onset}
\end{figure}

The same experiment family contains an operating point near the onset of temporal-route threshold
crossing. Its first small set of independent training runs suggested a split outcome, so a fresh
twenty-run experiment was frozen with those motivating runs excluded. Nine of twenty fresh runs
cross the registered probe $R^2$ threshold (Figure~\ref{fig:stochastic-onset}):
\begin{equation}
  \widehat{q}=\frac{9}{20}=0.45,
  \qquad
  \mathrm{CP}_{95\%}=[0.2306,0.6847].
\end{equation}
$\widehat{q}$ is the fraction of independent training runs that cross the threshold;
$\mathrm{CP}_{95\%}$ is the $95\%$ Clopper--Pearson interval for that rate.
A pre-registered empty band around the threshold contains no run. At the lower training budget,
zero of twenty runs cross. The runs that install also have better held-out cross-entropy in the
registered direction (one-sided Mann--Whitney $p=0.0015$), but this is a secondary association and
does not predict which run installs.

The result rules out a deterministic selector that depends only on the frozen architecture, data
distribution, schedule, and budget. It does not establish a thermodynamic phase transition, prove
that there are exactly two optimization basins, or identify an initialization statistic that predicts
the outcome. The estimand is a threshold-exceedance probability over independent training runs.

These experiments keep three statements distinct. A route can identify a
variable without clearing the admission threshold. It can clear the threshold when trained alone yet
lose assignment under competition. Near the onset of threshold crossing, a fixed experimental
configuration can put different independent training runs on different sides of that threshold.
Section~\ref{sec:theta4-reach-admit} is that first case on a different environment: a
temporal statistic that is identifiable from the input and not admitted under next-token
prediction. It is not a two-route competition, and it is not evidence about which eligible route
carries an admitted direction.

\section{Related work}
\label{sec:related}

\paragraph{Task-relative world models.}
Value equivalence replaces observation fidelity with agreement on a set of value functions
\citep{grimm2020,grimm2021}. Value-aware model learning and related objectives similarly weight model
errors by their effect on value \citep{farahmand2017,farahmand2018,abachi2020}; weighting the model
loss by the gradient of one value function \citep{voelcker2022vagram} is supervision at the rank-one
corner. One-step accuracy separates from control performance once predictions are composed over a
rollout \citep{talvitie2014model,lambert2020}. State abstraction, bisimulation, predictive-state
representations, and successor features all define model content relative to downstream distinctions
or predictions \citep{littman2001,singh2004,givan2003,ferns2004,dayan1993,barreto2017}. Low-rank
MDPs make a minimal model size a rank of the dynamics \citep{agarwal2020flambe}; closure rank is the
query-relative counterpart. Our closure object belongs to this lineage. The new contribution is the
separation between whether a closure direction is admitted and which of several training routes that
can identify it actually carries it.

Auxiliary-task work studies how reconstruction, self-prediction, and task signals shape learned
representations \citep{jaderberg2017,voelcker2024}. Auxiliary predictions shape value features
through the spectrum of the transition operator \citep{lyle2021effect}; supervision rank turns that
shaping into a count, verified direction by direction against a known closure. Task-informed
abstractions and denoised world models explicitly separate task-relevant structure from distractors
\citep{fu2021,wang2022}. Those works compare objectives or impose a factorization. We place multiple
routes to the same variable inside one model, measure the extra training loss when each route is
removed, and identify which route carries the variable with route-specific interventions.
Concurrently, \citet{halvagal2026task} show that value-based and policy-gradient agents acquire
different representational invariances under identical tasks; our results locate the analogous
route-dependence in world models, against an environment with known required coordinates, and
separate reachability from admission and assignment.

A second lineage derives the sufficient state for one committed query family
\citep{subramanian2022approximate,huang2022action,efroni2022provably,ni2024bridging}. Concurrent
work constructs or certifies task-sufficient models
\citep{feng2026tasksufficient,zheng2026specialist,richens2025general}. Those accounts fix the query
family or the construction; none separates admission from assignment. The longer lineage notes are
in Appendix~\ref{app:lineage}.

\paragraph{Feature competition and shortcuts.}
Shortcut learning describes reliance on predictive features that fail under distribution shift
\citep{geirhos2020}. Simplicity-bias constructions show that an easier feature can suppress a more
complex predictive feature, including conditions in which the harder feature is learned after the
simple feature is removed \citep{shah2020}. Feature preference also tracks availability in untrained
networks, and availability can override predictivity \citep{hermann2020,hermann2024}. These works own
the positive availability account and the idea of testing a suppressed feature in isolation.

Our assignment result asks a narrower forward question. It measures the extra training loss when the
unchosen route is removed, commits the predicted winner, certifies both routes at the same operating
point, and observes that the extra-loss prediction fails. Gradient starvation is a plausible
candidate mechanism for the outcome \citep{pezeshki2021}, but our experiments do not identify it.
Gradient starvation derives feature competition from coupled learning dynamics; extra training loss
was measured independently of those dynamics here.

Predictive objectives can themselves prefer temporally persistent nuisance variables because those
variables are easy to predict \citep{gulati2026}. That result concerns signal and nuisance under one
objective and confirms its mechanism intervention. Our construction supplies two routes to the same
task variable and falsifies the prediction that extra training loss selects the carrier. The two findings are complementary:
training signals shape representation, but extra training loss is not sufficient to infer which
of several identifying routes will carry a variable.

\paragraph{World-model measurement.}
Dreamer-family agents rely strongly on unsupervised world-model learning signals
\citep{hafner2020,hafner2021,hafner2023}. Whole-objective ablations show that those signals affect
return, but they do not identify the carrier of an individual latent direction. Our
frozen, low-capacity probe and same-model ablation table operate at that level. This measurement
choice also avoids changing the representation through adaptive readout \citep{interno2026}.
Effective rank has a separate life in deep RL as a training pathology: bootstrapped value learning
can collapse feature rank \citep{kumar2021implicit,kumar2022dr3}, and the association between rank
and return does not survive hyperparameter sweeps \citep{gulcehre2022empirical}. Installed rank is a
different quantity: a count of closure directions recovered above a registered threshold. In the
tested regime its value is predicted by supervision rank, not diagnosed from training dynamics.

The learned content studied here remains distinct from transition fidelity. Operator-based
diagnostics ask whether a learned transition advances a function family correctly
\citep{vakalis2026operator}; a model can capture a task-relevant coordinate and still propagate it
incorrectly. Admission and assignment are therefore necessary parts of a representation audit, not
a complete certificate for planning.

\section{Discussion}
\label{sec:discussion}

\paragraph{A scalar loss can contain high-rank supervision.}
The overall loss is scalar in every experiment. Rank belongs to the target covariance or, more
generally, to the independent prediction constraints presented over the data distribution. Counting
output units as though they were necessarily independent is therefore a mistake, as is treating
every scalar reward as rank one. In the linear target family the rank is exact and measurable. For
nonlinear rewards, part of the scientific problem is specifying what is being ranked: local
covectors, nonlinear sufficient statistics, or linearly usable content.

\paragraph{Closure states the requirement; routes explain acquisition.}
Closure alone does not predict a learned representation. It specifies the content sufficient for a
query family, while reach describes what each training channel can identify. The frame-observable,
recurrent-carry, filter-only, and supervision-only constructions show that several routes can admit
closure directions. Their ordering is architecture- and protocol-relative. We have not established a
universal progression from reconstruction to recurrence to filtering to supervision.

\paragraph{A reachable statistic can fail to enter the state.}
On the environment in Section~\ref{sec:theta4-reach-admit}, $\theta_4$ is identifiable from the training
input and is not recoverable from the trained state. Next-token cross-entropy prices accumulating it
at a fraction of a percent of the total training loss. Adding a head that predicts $\theta_4$, with
the same model family, data, and step budget, restores a generalizing read, seed-bistably at the
default auxiliary-loss weight. The model class can represent the statistic; next-token prediction by
itself does not install it. That is a reachability versus
admission split, not an assignment result, and it is scoped to this environment.

\paragraph{Extra training loss is not a carrier identification.}
The extra training loss paid when a route's information is removed is the loss available to that
route if it is exploited. It does not include when the feature becomes available, how gradients
interact after another route begins to learn, whether early allocation is entrenched, or how
initialization aligns the representation with one channel. Those quantities may explain assignment,
but they cannot be inferred from the extra-loss comparison that failed here. This paper does not
answer why the direct route wins. It answers what the why cannot be: not the static extra-loss
comparison, and not any deterministic function of architecture, data, schedule, and budget.

\paragraph{A decode threshold is not an information-theoretic claim.}
An operational representation claim needs a bar. Without one, arbitrarily small decoder improvements
can be redescribed as installation. A bar also creates a gray region. The recurrent-stack competitor
illustrates it directly: linear probe recovery remains below threshold while held-out likelihood improves. Reporting
both prevents a thresholded instrument from being mistaken for an information-theoretic statement.
Section~\ref{sec:theta4-reach-admit} is the same distinction with a positive control: the statistic is in
the input, and a supervision head installs a generalizing read. The stochastic-boundary experiment further treats threshold crossing as a probability over
independent training runs rather than
assigning a deterministic label to the training configuration.

\paragraph{Limitations.}
The evidence comes from controlled plants and three compact world-model stacks. The controlled
observation process is a measurement device and supports no natural-image claim. The
filtering-identifiable coefficient in Section~\ref{sec:theta4-reach-admit} is exploratory mechanism
evidence on one environment and training protocol, obtained after a pre-registered passive
evaluation; it is not that evaluation, and at the
default auxiliary-loss weight the restored admission is seed-bistable. The strongest
competition over which route carries the variable is established on one plant and one DreamerV3-style
stack; the transformer/RSSM comparison transfers an extra-loss versus probe ordering on the same
recorded sequences but does not supply a second experiment in which each route carries when trained
alone. Supervision-rank equality is empirical for nonlinear heads, and its
linear ceiling uses the measured premise that unsupported directions stay at the reconstruction null.
The installed-rank instrument is a frozen
linear probe with a registered threshold; nonlinear or distributed codes below that threshold may
still carry information. Finally, neither the extra-loss negative nor the seed-level split identifies the
optimization mechanism selecting a carrier.

These limitations define concrete tests, stated as hypotheses in
Appendix~\ref{sec:synthesis}: whether subspaces admitted without the task head and subspaces
constrained by supervision combine by their union rather than by target count; whether the extra-loss
versus carrier dissociation survives a hidden-oscillator geometry; and how the onset of threshold
crossing, across independent training runs, changes with unique training content. Each can fail
without changing the distinction between reachability, admission, and assignment.

\section{Conclusion}
\label{sec:conclusion}

What a world model represents is not decided by a scalar objective in one step. A task-relevant
direction must be identifiable from the information available to training. Directions left behind by
reconstruction, recurrence, or filtering can be recruited by explicit supervision, and in our controlled construction their
installed rank tracks the effective rank of the supervised target family. The rank-one value result
is one instance of this admission law, not a consequence of scalar loss. A filtering-identifiable
coefficient can still fail to enter a state trained only for next-token prediction when accumulating
it is worth a fraction of a percent of that loss; adding a head that predicts the coefficient
restores it in the same model family.

When routes overlap, admission ceases to identify the carrier. Reversing which route is more costly
to remove does not reverse assignment in the competition where each route carries when trained alone.
The broader extra-loss versus probe ordering transfers
across two architectures on the same recorded sequences, and the onset of threshold crossing is
stochastic across independent training runs. Reachability governs eligibility; independent supervision governs
admission of leftover directions in the tested regime; and when several routes can identify a variable, extra training loss
does not name the carrier. A different failure needs a different remedy: unreachable information
cannot be recovered by stronger supervision, a reachable statistic can still fail to enter the state
if next-token prediction does not pay to accumulate it, a rank-deficient target cannot be repaired by weighting
the same scalar more heavily, and extra training loss does not identify the carrier.

\bibliographystyle{abbrvnat}
\bibliography{references}

\appendix
\section{Additional measurement details}
\label{app:measurement}

\subsection{Admission probes and controls}

All registered admission reads use frozen representations and held-out ridge probes. The main query head
and the corresponding probe read $z$ alone; $h$-only reads check that the query is not being
satisfied through the recurrent state. Probe
regularization is selected without access to the held-out evaluation targets. Each supervision-rank
setting reports the coordinatewise gaps, full-closure $R^2$, installed count, reconstruction-only
and shuffle baselines, and how the installed count moves if the threshold is varied.

The auxiliary-head staircase is replicated at every rank, with three seeds at the two interior ranks
where threshold sensitivity is most consequential. The value-head staircase is claimed on the
three-seed mean at rank two together with the rotation-invariant full-closure $R^2$. The borderline
closure coordinate that the target does not supervise, in one rank-two seed, is retained in all
per-seed displays rather than deleted as an
outlier.

\subsection{Linear constraint rank, empirical installation, and capacity}
\label{app:rank-ceiling}

Let $Y=U^\top L$ and suppose a linear head predicts $Y$ from latent $z$. Then
\begin{equation}
  \Sigma_{Y,z}=U^\top\Sigma_{L,z},
  \qquad
  \operatorname{rank}(\Sigma_{Y,z})\leq \rsup.
\label{eq:crosscov-bound}
\end{equation}
$\Sigma_{Y,z}$ is the cross-covariance of the target with the latent, $\Sigma_{L,z}$ is the
cross-covariance of the closure coordinates with the latent, and $\rsup$ is supervision rank.
Reduced-rank regression identifies the canonical directions selected by the best rank-constrained
linear predictor \citep{anderson1951,eckart1936,brillinger1969,izenman1975}. The learned encoder need
not be the unique population optimum off those supervised directions. Consequently, an installed-rank
ceiling also requires the empirical premise that unsupported directions remain at the null.
Reconstruction-only, unsupervised-column, and shuffled-target controls test that premise in the
present construction.

Capacity limits attainment but does not set the observed plateau once capacity is sufficient. On a
continuous-latent calibration, installed rank increases with latent width and then plateaus at the
supervision rank through latents several times wider than the closure. This capacity-independent
calibration is synthetic; a corresponding trained high-dimensional-observation experiment installed
the closure on its training distribution but generalized poorly to fresh rollouts, and we do not use
it as evidence of pixel-scale generality.

\section{Validity of the probe that identifies the carrier}
\label{app:carriage}

The probe that identifies the carrier is an intervention table on a frozen representation. Each
route is removed at read time on the same trained model. A single-carrier label requires a selective
drop when the proposed route is removed, preservation when the other route is removed, a
no-information floor when both are removed, a placebo substitution that should not matter, and a
live probe. A training control that removes one route establishes that each candidate route can
carry at the operating
point. The readout is low capacity and does not adapt the world model.

Cross-episode substitutions are checked for in-support inputs and for how far the substituted stream
moves, because removing information can
also create an out-of-distribution input. The probe returns an instrument-failure state when a
substitution has excessive effect on the other route or when a positive control is absent. It never converts missing
evidence into a route label.

\section{Scope of the cross-architecture transfer}

The recurrent comparison uses the transformer's recorded sequences. It shows
that the extra-loss versus probe ordering persists under an architectural transfer. It is not independent data,
does not include the transformer's longer-budget stochastic operating point, and does not establish that
each route carries under passive training on the RSSM. The supervised controls establish that the
shared model body can represent the variable at the two
ends of the noise knob; interior settings are bracketed rather than directly supervised. Dependence is assessed
at the independent-training-run level. Below-threshold temporal-route reads are accompanied by cross-entropy and
nonlinear-probe secondaries.

\section{Data-regime boundary}
\label{app:data-regime}

The filtering-route data sweep uses views per underlying frame as its primary count: how often
training revisits each distinct observation. It also
reports total unique content alongside it. Installation remains above the reconstruction-only
baseline through the poorest setting three independent runs can still decide,
giving the one-sided boundary $D_{\mathrm{crit}}>410$ views per frame: at $40$ unique episodes the
parameter still installs. At $4$, $10$, and $20$ unique episodes,
three runs cannot resolve whether the parameter installed: the reads lie in the
predefined ambiguous band, and the memorization statistic does not reach the signature of a model
that has only stored a small working set. These
settings are reported as undecided rather than negative. Lengthening trajectories cannot repair
uncertainty across independent training runs in the raw decoder; future refinement requires more
runs or denser reads.

\section{Filtering-identifiable coefficient: construction and scope}
\label{app:theta4}

The $\theta_4$ measurements in Section~\ref{sec:theta4-reach-admit} use a categorical RSSM trained
with next-token cross-entropy on a synthetic environment in which one episode-level autoregressive
coefficient is carried only in the lag-1 second moment of one observation channel. The competing
single-frame render of that coefficient is removed during training. Two versions are reported: a
four-parameter environment and an environment that contains only $\theta_4$, both with the same
temporal channel. The supervision control uses the $\theta_4$-only environment, the same
architecture and step budget as the passive $\theta_4$-only runs, and a linear head on the
concatenated posterior state with mean-squared error on $\theta_4$.

The incentive number is an exact-law gap, not a fitted predictor. On the relevant channel tokens,
the generator supplies a current-frame-only predictor (the coefficient unknown) and a filter that
integrates the exact posterior over the coefficient from the quantized history. The difference is
$0.041$--$0.045$ nats. Because that channel is one of seven fields, the share of total
cross-entropy is that gap divided by seven, then by the sequence cross-entropy, giving
$0.32$--$0.35\%$ on every measured passive model. A matched channel with a fixed
autoregressive coefficient and no episode-level latent sits at the one-step exact law, confirming
that the model can match an exact one-step predictor on a channel that has no episode coefficient.

A pre-registered passive evaluation on this environment failed to recover $\theta_4$ from the
state. The numbers in Section~\ref{sec:theta4-reach-admit} are a later exploratory mechanism
analysis, not that evaluation. The claim is about this environment, not a general fact about world
models. Failure under this reader and threshold is not proof of zero information. The result
is not assignment evidence: only one identifying route was available.

One passive $\theta_4$-only seed carries a persistent, direction-stable, seed-specific weak
$\theta_4$ trace, about $0.8\%$ of a paired-perturbation state difference, below every bar used in
the pre-registered evaluation. It is not part of the headline result.

\section{Observation covariance is not closure}
\label{app:cov-closure}

\begin{figure}[h]
  \centering
  \includegraphics[width=.92\linewidth]{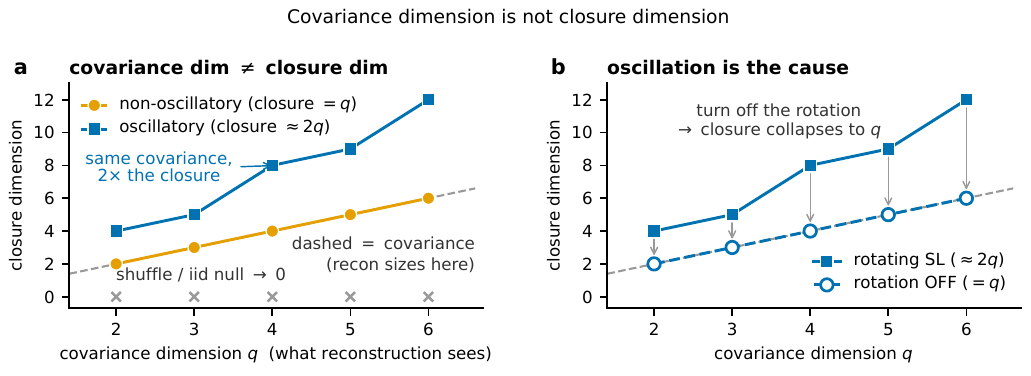}
  \caption{\textbf{Observation dimension and predictive closure are different objects.} An
  oscillatory signal occupies one covariance direction but requires a two-dimensional predictive
  state. Closure defines the content whose admission and assignment we study.}
  \label{fig:cov-closure}
\end{figure}

\section{Span geometry as a hypothesis}
\label{sec:synthesis}

The admission and assignment results suggest a single geometric experiment. Let
$P\subseteq\closure$ denote the subspace admitted by reconstruction, recurrence, or filtering and
$O\subseteq\closure$ the subspace directly constrained by an explicit query, reward, or value target. In a regime where
latent capacity is not the binding constraint, the prediction we test is
\begin{equation}
  k_{\mathrm{installed}} \approx \dim(P\vee O),
\label{eq:lattice-prediction}
\end{equation}
where $k_{\mathrm{installed}}$ is the number of admitted closure directions and $P\vee O$ is the span
of the two subspaces. If $j=\dim(P)$ and the supervised target has rank $\rsup$, generic position
gives $\min(j+\rsup,k)$, with $k$ the dimension of $\closure$. That count overestimates whenever the two
subspaces overlap. A one-dimensional overlap of $P$ and $O$ therefore distinguishes a genuine subspace law
from the simpler story that each additional target coordinate adds another latent direction.

Equation~\ref{eq:lattice-prediction} is a hypothesis, not a theorem or a completed empirical result.
It requires enough capacity in $z$, stable admission thresholds, a rotation-invariant read of each
subspace, and a principal-angle estimate of $P\cap O$. It can fail through displacement, nonlinear
spillover, optimization noise, or a subspace admitted without the task head that changes after that
head is added.
The last possibility is a transport failure: showing that a direction installs without the task head
does not automatically transfer to a jointly trained model.

The same construction can separate admission from assignment. A direction in $P\cap O$ is reachable
from reconstruction, recurrence, or filtering and also from an explicit query or value head. Changing which of
those routes is more costly to remove, while holding the
direction admitted, tests whether extra training loss changes admission at all; route-specific interventions are
then needed to determine assignment. Directions in $O\setminus P$ test supervision rank on
directions not already admitted without the task head,
while directions in $P\setminus O$ verify that content admitted without the task head survives target shuffling.

An independent-geometry version replaces the known-input temporal route with a hidden oscillator
whose single-frame marginal is exactly independent of the episode parameter. The parameter is available
only through temporal autocorrelation, while a separate reconstruction channel supplies a direct route. A
reversal in which channel is more costly to remove then provides a two-sided test of whether the
extra-loss versus carrier dissociation extends beyond the original plant. A reversal would establish a geometry interaction;
a repeated non-reversal with each route shown to carry when trained alone would support a broader empirical law. Failure to
make each route carry when trained alone would be a feasibility boundary, not assignment evidence.

Supervision rank predicts how many residual directions the explicit target can recruit. Span
geometry predicts how those directions add to passive content. Route interventions ask how an
overlapping direction is carried. The three questions share a closure reference but have different
estimands and can receive different answers.

\section{Lineage notes}
\label{app:lineage}

\paragraph{Sufficient states for a fixed query family.}
A second lineage derives the sufficient state for one committed query family, and each member reads
as a fixed instance of closure. An approximate information state must predict the instantaneous
reward and its own successor \citep{subramanian2022approximate}: the second requirement is closure
under the dynamics; the first fixes a rank-one query family. Action-sufficient state representations
characterize the minimal set of state dimensions sufficient for policy learning
\citep{huang2022action}: that minimal sufficient state is the closure specialized to the return
query, effective rank generalizes their minimal set of state dimensions, and supervision rank gives
the size of the installed set as a function of the prediction constraints in our controlled regime.
Bisimulation-metric encoders keep exactly the distinctions one reward can make
\citep{zhang2021learning}. Multistep inverse models provably recover the control-endogenous state,
which is the closure of the controllability queries \citep{efroni2022provably,lamb2023guaranteed},
and provably miss it without a latent forward constraint \citep{levine2024multistep}; the repair is
a closure condition. A representation can also maximize several mutual-information objectives while
remaining unable to represent the optimal policy \citep{rakelly2021which}: those objectives can be
optimized even though a direction the task needs is never admitted. Self-predictive abstraction unifies many of
these conditions \citep{ni2024bridging}. Reward, return, optimal action, controllability: each work
fixes the query family in advance. Closure is the same object with the query family left free, and
admission and assignment ask how its directions arrive in a learned latent.

\paragraph{Task-sufficient world models.}
Concurrent work arrives at task-relative model content from the construction side.
\citet{feng2026tasksufficient} learn world models that are minimal, sufficient, and task-specific by
coupling agentic exploration with structured representation learning, and \citet{zheng2026specialist}
give identifiability conditions for the task-relevant subspace of a generalist representation. A
position paper organizes latent-state design by what the state must be sufficient for
\citep{kim2026latent}; offline diagnostics predict closed-loop performance from the observability of
one task's reward \citep{smolyanskiy2026predicting}; Bayesian experiment design asks which
interventions identify a mechanistic world model at all \citep{murphy2026mda}.
\citet{richens2025general} prove that an agent generalizing across multi-step goal-directed tasks
must have learned a predictive model whose required accuracy grows with the goal family and depth.
That theorem is the closest formal cousin to ``quality is relative to a query family and horizon.''
These works
acquire, organize, or certify task-sufficient models; none counts the closure directions a query
family requires, and none separates admission from assignment.

\section{Evidence map}

\begin{table}[h]
\centering
\caption{Claim grades used in the paper. ``Exact'' refers to the controlled construction, not a
universal property of learned world models.}
\begin{tabular}{L{.31\linewidth}L{.20\linewidth}L{.39\linewidth}}
\toprule
Claim & Grade & Scope \\
\midrule
Admission is contained in joint reach & theorem & exhaustive registered training information set \\
Linear target constraint rank is at most $\rsup$ & construction-exact & linear target and second-moment covariances \\
Installed rank equals $\rsup$ & empirical law & controlled latent-recovery regime with null stability \\
Demand does not select the carrier & empirical counterexample & DreamerV3 competition where each route carries alone, at the tested settings \\
Extra-loss versus probe ordering crosses architectures & transferred result & same recorded sequences; transformer and RSSM \\
Boundary assignment is stochastic & empirical boundary & registered threshold and fresh independent training runs \\
Reachable $\theta_4$ is not admitted under passive CE & exploratory mechanism & one environment and protocol; after a pre-registered evaluation; not assignment \\
$k_{\mathrm{installed}}\approx\dim(P\vee O)$ & hypothesis & abundant capacity; overlap instrument required \\
\bottomrule
\end{tabular}
\end{table}

\end{document}